%% file: main.tex
\documentclass[10pt,twocolumn,letterpaper]{article}

\usepackage[pagenumbers]{cvpr} 

\usepackage{graphicx}
\usepackage{amsmath}
\usepackage{amssymb}
\usepackage{booktabs}
\usepackage[pagebackref,breaklinks,colorlinks]{hyperref}
\usepackage[capitalize]{cleveref}
\crefname{section}{Sec.}{Secs.}
\Crefname{section}{Section}{Sections}
\Crefname{table}{Table}{Tables}
\crefname{table}{Tab.}{Tabs.}

\def\cvprPaperID{*****} 
\def\confName{CVPR}
\def\confYear{2022}

\input{others/preamble}
\begin{document}
\input{others/metadata}
\maketitle
\thispagestyle{empty}
\input{sections/0_abstract}
\input{sections/1_introduction}

\input{sections/2_related}
\input{sections/3_rethinking}
\input{sections/4_experiments}
\input{sections/5_summary}

{
    \small
    \bibliographystyle{ieee_fullname}
    \bibliography{others/default}
}

\clearpage
\input{sections/X_supplementary}

\end{document}

%% file: others/preamble.tex
\usepackage{overpic}
\usepackage{enumitem} 
\usepackage{overpic} 
\usepackage{color}

\definecolor{turquoise}{cmyk}{0.65,0,0.1,0.3}
\definecolor{purple}{rgb}{0.65,0,0.65}
\definecolor{dark_green}{rgb}{0, 0.5, 0}
\definecolor{orange}{rgb}{0.8, 0.6, 0.2}
\definecolor{red}{rgb}{0.8, 0.2, 0.2}
\definecolor{darkred}{rgb}{0.6, 0.1, 0.05}
\definecolor{blueish}{rgb}{0.0, 0.3, .6}
\definecolor{light_gray}{rgb}{0.7, 0.7, .7}
\definecolor{pink}{rgb}{1, 0, 1}
\definecolor{greyblue}{rgb}{0.25, 0.25, 1}

\usepackage{others/mmstyles} 

\usepackage{blindtext}

\renewcommand{\paragraph}[1]{\vspace{1em}\noindent\textbf{#1}.}

\usepackage{adjustbox}
\usepackage{colortbl}
\usepackage{diagbox}
\definecolor{highlight}{cmyk}{0,0,0,0.29}

%% file: others/metadata.tex
\title{Rethinking Unsupervised Domain Adaptation for Semantic Segmentation}

\author{
    Zhijie Wang\textsuperscript{\rm 1},
    Masanori Suganuma\textsuperscript{\rm 1,2},
    Takayuki Okatani\textsuperscript{\rm 1,2} \\
    \textsuperscript{\rm 1} Graduate School of Information Sciences, Tohoku University \\
    \textsuperscript{\rm 2} RIKEN Center for AIP \\
    {\tt\small \{zhijie, suganuma, okatani\}@vision.is.tohoku.ac.jp} \\
}

%% file: sections/0_abstract.tex
\begin{abstract}
Unsupervised domain adaptation (UDA) adapts a model trained on one domain (called source) to a novel domain (called target) using only unlabeled data. Due to its high annotation cost, researchers have developed many UDA methods for semantic segmentation, which assume no labeled sample is available in the target domain. We question the practicality of this assumption for two reasons. First, after training a model with a UDA method, we must somehow verify the model before deployment. Second, UDA methods have at least a few hyper-parameters that need to be determined. The surest solution to these is to evaluate the model using validation data, i.e., a certain amount of labeled target-domain samples. This question about the basic assumption of UDA leads us to rethink UDA from a data-centric point of view. Specifically, we assume we have access to a minimum level of labeled data. Then, we ask how much is necessary to find good hyper-parameters of existing UDA methods. We then consider what if we use the same data for supervised training of the same model, e.g., finetuning. We conducted experiments to answer these questions with popular scenarios, \{GTA5, SYNTHIA\}$\rightarrow$Cityscapes. We found that i) choosing good hyper-parameters needs only a few labeled images for some UDA methods whereas a lot more for others; and ii) simple finetuning works surprisingly well; it outperforms many UDA methods if only several dozens of labeled images are available.
\end{abstract}

%% file: sections/1_introduction.tex
\section{Introduction}
\label{sec:intro}

Domain adaptation is the problem of adapting a model pre-trained on a domain called the {\em source domain} to a different domain called the {\em target domain}. Unsupervised domain adaptation (UDA) is the most widely studied variant. UDA first trains a model in a supervised manner using the labeled source-domain samples and then adapts the model to work well on the target domain using {\em unlabeled} target-domain samples. Many studies have been conducted to develop UDA methods for {\color{black} image classification \cite{ghifary2016deep}, object detection \cite{zheng2020cross}, semantic segmentation \cite{adaptsegnet,fada}}, etc. This study concentrates on semantic segmentation, for which various methods have been developed, such as those based on adversarial training \cite{adaptsegnet,advent}, self-training \cite{cbst,crst}, and their combinations \cite{iast}. 

Almost all these studies assume that {\em no labeled samples are available} in the target domain. This study first questions this assumption. We believe it is practically unrealistic since we usually need labeled samples in the target domain for a few reasons. First, we need to check the performance of models before deployment. The most standard way is to evaluate their performance using a certain amount, if not many, of labeled target-domain samples. Second, we need to choose the hyper-parameter every UDA method must have. Despite several attempts in the literature to do it without labeled target-domain samples, the surest way is arguably to use them. 

Why do the previous UDA studies not consider even a slight deviation from this assumption? To the authors' knowledge, no study provides an explanation or a realistic, practical scenario for this setting. Previous studies concentrate on methodologies but do not seem to pay much attention to data usage. While it is typical for academic research, the attitude may not properly handle some of the real-world demands. It is sometimes most important for practitioners to decide how to utilize data.

Following this practitioner's perspective, we locate ourselves in a data-centric position. We first assume that labeled target-domain samples are available; considering the high annotation cost that motivates UDA, we assume a small number of labeled samples to be available. Then, we consider how we should use them to maximize inference accuracy. Alternatively, we consider how much data are necessary to achieve a target we set. 

We consider two natural usages of labeled target-domain samples and compare their effectiveness. One is to use them for tuning hyper-parameters of UDA methods. Any UDA methods have one or more hyper-parameters, and it is reasonable to choose them in this way. Researchers have not paid much attention to the importance of tuning hyper-parameters with UDA methods until recently. Saito et al. pointed out the issue with existing studies and proposed a zero-shot approach, i.e., tuning UDA methods' hyper-parameters without using a labeled target-domain sample \cite{saito2021tune}. While such methods, including more general-purpose methods \cite{sugiyama2007covariate,you2019towards}, are attractive, they have their issues and cannot be a complete solution. 

The other is to use labeled target-domain samples for supervised training. While there may be several candidates, this study considers the most straightforward one, i.e., using them to finetune the pre-trained model. We consider how it performs compared with UDA methods for different amounts of labeled samples. Note that our finetuning method does not use unlabeled target-domain samples, which UDA methods use. While using them additionally is expected to be beneficial, we leave it to a future study; we concentrate on the simple finetuning to reveal the bottom-line impact of using labeled samples. Although the setting of using labeled and unlabeled samples jointly is considered by semi-supervised DA (SSDA), there is no well-established method. It should also be noted that the previous SSDA studies assume that more than 100 labeled samples (i.e., images) are available, whereas we consider a different range from one labeled sample to at most several dozens.






In this study, we limit our attention to semantic segmentation. The reason is multi-folds. One is the high demands for domain adaptation with the task due to its high annotation cost. Another is the characteristics of the annotation. The manual annotation is usually performed image-wise, which makes the annotation costly. Thus, we mainly consider cases where a limited number of images can be annotated. On the other hand, as an image consists of many pixels, a single annotated image is equivalent to having many labeled samples in the case of image classification.

%% file: sections/2_related.tex
\section{Related Work}

\subsection{Unsupervised Domain Adaptation}

This paper concentrates on unsupervised domain adaptation (UDA) for semantic segmentation, although we believe our argument holds for other tasks. There are two approaches to UDA: adversarial training and self-training. The former primarily aims to reduce the domain gap through adversarial training in feature space~\cite{fada}, input space~\cite{gong2019dlow,lee2018diverse}, or output space~\cite{adaptsegnet}. Self-training creates pseudo labels for target domain samples and uses them to finetune a pretrained model~\cite{PFAN_2019_CVPR}. To create more accurate pseudo labels, CBST~\cite{cbst} and CRST~\cite{crst} use class-balanced self-training and confidence-regularized self-training, respectively. Several attempts have recently been made to increase performance by integrating adversarial training with self-training. BLF~\cite{li2019bidirectional} use pseudo labels without any filtering. To cope with erroneous labels, AdaptMR~\cite{zheng2020unsupervised} filters pseudo labels and ignores the filtered-out pixels; it could discard useful information. IAST~\cite{iast} also filters pseudo-labels for self-training and employs entropy minimization to optimize the filtered-out pixels.

Previous studies of UDA, including the above, share a common problem in the experimental evaluation: there is no clear separation between validation and test data. While often not explicitly stated, they select their hyperparameters and evaluate the model's performance on the same data. In this paper, we examine the UDA methods' sensitivity to the hyperparameters and also consider how many labeled target-domain samples are necessary to choose the hyperparameters. 


\subsection{Selecting Hyper-parameter for UDA Methods}

Several studies consider choosing hyper-parameters of UDA methods without labeled target-domain samples. A simple way is to use the source-domain risk as a proxy of the target-domain risk~\cite{ganin2015unsupervised}. Sugiyama et al.~\cite{sugiyama2007covariate} and You et al.~\cite{you2019towards} propose methods that weight source-domain risks according to the similarity between target and source samples.
However, if there is a large domain gap,  the two risks deviate from each other and these methods do not work. Morerio et al.~\cite{morerio2017minimal} propose to use the entropy of classifiers as an unsupervised criterion for hyper-parameters selection. However, this method may fail to find good hyper-parameters when target samples are wrongly classified with high confidence, which often occurs in practice. Saito et al.~\cite{saito2021tune} propose to use the soft neighborhood density (SND) to address the instability of the entropy-based validation. SND requires the users to specify temperature for softening softmax probabilities, which needs to be chosen in some ways. 


\subsection{Finetuning for Few-shot Learning}

Finetuning has recently been found to be an effective method for few-shot learning, i.e., training a model to classify novel classes from only their few-shot training samples. Previously, meta-learning-based approaches were believed to be the best for the problem. However, several recent studies have discovered that simple finetuning on the novel class classification using provided few-shot examples outperforms meta-learning-based methods. The key is to train a model on the flat classification of all base classes using their training samples in advance. Wang et al.~\cite{wang2020few} have shown that a simple method for few-shot object detection works well, which finetunes only the last layer of an object detector on a class-balanced dataset. Tian et al.~\cite{tian2020rethinking} present a method for few-shot image classification, which trains a feature extraction network and uses it to map the query and support images into the feature space. For few-shot image segmentation, Boudiaf et al.~\cite{repri} propose RePRI, which finetunes the classifier using the support samples during the inference time to improve inference accuracy.
These studies and our interest share the interest in finetuning models with small amount of training samples. 
However, they (implicitly) assume that the base and novel classes belong to the same domain, which differs from domain adaptation. 

\subsection{Semi-supervised Domain Adaptation}

Semi-supervised domain adaptation (SSDA) relaxes the assumption of UDA. Namely, SSDA attempts to reduce domain gap using some amount of labeled target-domain samples in addition to unlabeled ones. We consider a similar setting where both unlabeled and labeled samples are available. However, the previous SSDA studies deal with only the setting that relatively many labeled samples ($\geq 100$) are available. In this study, we consider settings with fewer labeled samples (i.e., from one to a few dozens). The difference may change the problem's fundamental nature and lead to a significant difference in real-world applicability. It is also noted that the existing SSDA studies share the same potential issues with the experimental design as UDA, i.e., no clear separation between validation and test data. 

The representative studies of SSDA are as follows. Wang et al.~\cite{wang2020alleviating} propose a dual-level adversarial training; the first is an image-level adaptation, which aligns the global feature distributions, and the second is a semantic-level adaptation, which eases the semantic-level misalignment between the source and target features of the same class. Chen et al.~\cite{chen2021semi} mix the source-domain data and the target-domain data at a region-level and a sample-level to produce the domain-mixed data, training teacher models. Then they train a student model by knowledge distillation from the teacher models. 

%% file: sections/3_rethinking.tex
\section{Rethinking UDA from Data-centric Perspective}

\subsection{Realism of the Assumption of UDA}
The existing UDA studies assume that no labeled target-domain sample is available at training  (i.e., adaptation) time. Then, they aim to achieve as good performance as possible by performing adversarial training \cite{adaptsegnet,advent}, pseudo labeling \cite{cbst,crst}, their combination \cite{iast}, etc. However, while it simplifies the problem, {\em is this assumption realistic from a practical point of view?} Few studies have paid attention to the practicality of the assumption. To the authors' knowledge, no study provides a realistic, practical scenario of the setting. 

We usually need a validation step to check models' performance before deployment. Not many practitioners will be comfortable without this step. The most natural way to do it is to measure the performance using some amount of labeled samples. Note that it may not be impossible without a labeled sample of the learned task itself, e.g., using some proxy task (e.g., a downstream task utilizing segmentation results) as evidence for the model's performance. However, this will be feasible only in a limited number of cases. 

Assuming we have, if only a tiny amount, labeled target-domain samples, why do we not spare some of them to improve the model's performance? (We will use the rest for validation.) So, we explore this idea. Namely, we assume we have a small amount of labeled target-domain samples, slightly more than the minimum requirement for the model's validation. We then ask: what is the best way to use them to achieve maximum performance.

 






\subsection{Choosing Hyper-parameters of UDA Methods }

One natural use of labeled target-domain samples is to use them to tune the hyper-parameters of a UDA method. Any UDA method has hyper-parameters, which need to be selected somehow. Unfortunately, most previous studies on UDA for semantic segmentation \cite{advent,intrada,fada,iast} do not handle it properly, as discussed in Sec.~\ref{sec:issue_cityscape}. 
Different UDA methods will show different sensitivities to the choice of their hyper-parameters. Some methods have many hyper-parameters, and others have only a few. So, we first consider the following research questions regarding the existing methods for UDA for semantic segmentation.

\medskip
\noindent
{\bf RQ1}: {\em How sensitive is their performance to the choice of hyper-parameters?}

\medskip
\noindent
{\bf RQ2}: {\em How many labeled samples are necessary to choose the optimal or nearly optimal hyper-parameters?
}

\subsection{An Alternative Approach: Finetuning}
\label{sec:finetuning}

Another way of using labeled target-domain samples is to use them for supervised training. UDA methods adapt a model pre-trained on the source domain in a super-vised fasihon. We consider finetuning the same pre-trained model using the available labeled target-domain samples. It is arguably the most straightforward approach to transfer a model across different domains. Then, we ask the following questions:

\medskip
\noindent
{\bf RQ3}: {\em How effective is it to do finetuning the model using a small amount of labeled target-domain samples?}


\medskip
\noindent
{\bf RQ4}: {\em Which performs better between existing UDA methods and the finetuning method in an equal condition in the number of available labeled samples?} 


\medskip
Note that our finetuning method does not use the unlabeled target-domain samples that UDA methods use. Thus, the comparison of the two is not fair. We concentrate on the simple finetuning method for simplicity and two more reasons. Firstly, using additional unlabeled target-domain samples will undoubtedly lead to improvement. (The question is how significant the improvement will be.) Thus, we first want to estimate the minimum impact of using labeled samples. Secondly, it is unclear how we can utilize the two types of samples; thus, we leave it a future study. The setting of using a combination of labeled and unlabeled samples is called semi-supervised domain adaptation (SSDA), for which several studies exist. However, the existing studies deal with only the case where relatively many labeled samples are available. The existing SSDA studies assume 100--500 labeled samples (i.e., images), whereas our setting assumes at most several dozens of labeled samples, following the motivation behind UDA. It is unclear if the method known to be effective for the former is effective for the latter.  



\subsection{Issues with the Experimental Design of UDA}
\label{sec:issue_cityscape}

As mentioned above, previous studies of UDA for semantic segmentation do not follow the standard machine learning procedure for the split of data. This is partly because they divert the datasets designed for the standard setting of supervised learning to domain adaptation. Most of them employ the Cityscapes dataset \cite{cityscapes} for the target-domain data; GTA5 \cite{gta5} and SYNTHIA \cite{synthia} are used for the source-domain data. The Cityscapes dataset provides the official training/validation/test splits. The  training and validation sets are publicly available, whereas the test set is not; only input images are available. To evaluate a method's performance one need to submit its result on these imaegs to the official server\footnote{\url{https://www.cityscapes-dataset.com/submit/}}.

Arguably due to this limitation, the previous studies of UDA use the same official data split; that is, they use the training set for adaptation and evaluate the methods' performance on the validation set. Thus, they do not use a separate set for validation; some use the same validation set for choosing hyper-parameters and others do not mention how to do it. 

In our experiments, we split the official validation set into customized validation and test sets to rectify the above issue.

%% file: sections/4_experiments.tex
\section{Experiments}


\subsection{Experimental Settings}


We consider the following six UDA methods for semantic segmentation: AdaptSegNet \cite{adaptsegnet}\footnote{\url{https://github.com/wasidennis/AdaptSegNet}}, AdvEnt \cite{advent}\footnote{\url{https://github.com/valeoai/ADVENT}}, CBST \cite{cbst}\footnote{\url{https://github.com/yzou2/CRST}}, FADA \cite{fada}\footnote{\url{https://github.com/JDAI-CV/FADA}}, IntraDA \cite{intrada},\footnote{\url{https://github.com/feipan664/IntraDA}} and IAST \cite{iast}\footnote{\url{https://github.com/Raykoooo/IAST}}. We used their public implementations, all of which employ DeepLabv2 \cite{deeplab} based on a ResNet-101 backbone. We also followed the original training procedures including data augmentation except FADA, for which we skipped the final self-distillation step for simplicity. We conducted all experiments using PyTorch \cite{pytorch} and Nvidia 2080Ti except IAST, for which we used Nvidia V100.


Following the previous studies, we choose the Cityscapes dataset \cite{cityscapes} for the target domain
and GTA5~\cite{gta5} and SYNTHIA~\cite{synthia} for the source domains. We consider two adaptation scenarios, i.e., GTA5 $\rightarrow$ Cityscapes and SYNTHIA $\rightarrow$ Cityscape.


To rectify the issue with the split of data in the previous studies' experiments explained in Sec.~\ref{sec:issue_cityscape},  we create and use a customized validation set $\mathcal{S}_\mathrm{val}$ and test set $\mathcal{S}_\mathrm{test}$ from the Cityscapes' official validation set. Specifically, we create $\mathcal{S}_\mathrm{val}$ and $\mathcal{S}_\mathrm{test}$ in the following way. We first choose 400 images to form $\mathcal{S}_\mathrm{test}$  (i.e., $\lvert\mathcal{S}_\mathrm{test}\rvert=400$) from the official validation set, which include 500 images. We then choose $\mathcal{S}_\mathrm{val}$ with different sizes in the range of $[1,100]$ from the remaining 100 images. 

We then use these splits to evaluate UDA methods and finetuning in the following way. For UDA methods, we use $\mathcal{S}_\mathrm{val}$ for selecting their hyper-parameters. For finetuning, 
we use $\mathcal{S}_\mathrm{val}$
as a `training' set despite its name. We use $\mathcal{S}_\mathrm{test}$ for the evaluation of the UDA methods and finetuning. To eliminate biases caused by a specific split, we iterate the above procedure randomly and create the following sequence:
\begin{equation}
(\mathcal{S}_\mathrm{val}^{(1)},\mathcal{S}_\mathrm{test}^{(1)}), (\mathcal{S}_\mathrm{val}^{(2)},\mathcal{S}_\mathrm{test}^{(2)}), \ldots, 
(\mathcal{S}_\mathrm{val}^{(N_\mathrm{trial})},\mathcal{S}_\mathrm{test}^{(N_\mathrm{trial})})
\label{eqn:split}
\end{equation}
and report the statistics (i.e., the average and variance) over $N_\mathrm{trial}$ trials. We set $N_\mathrm{trial}=30$ and use an identical sequence for all the methods.


\subsection{Results}

\subsubsection{Sensitivity of UDA Methods to Hyper-parameters} \label{ch:rethinking_hp}

We first consider RQ1 and examine the sensitivity of the above six UDA methods to the selection of their hyper-parameters. We consider adaptation from GTA5 to Cityscape here.


While each method has a different number of hyper-parameters, we consider only the primary ones for each for experimental efficiency. For IntraDA~\cite{intrada}, we consider $N_\mathrm{iter}$, the number of iterations to update the feature space, and $\lambda$, a threshold for the entropy over class probabilities to select the easy samples and hard samples in the target domain. {\color{black} For AdaptSegNet~\cite{adaptsegnet}, we consider $N_\mathrm{iter}$ and the weight $w$ of the auxiliary segmentation loss. For AdvEnt~\cite{advent}, we have $N_\mathrm{iter}$ and initial learning rate $lr$. For CBST~\cite{cbst}, we consider the initial proportion $p_0$ of the pseudo label pixels and the proportion change $\Delta p$ of every training round. For IAST~\cite{iast}, we consider the initial proportion $\alpha$ of the pseudo label pixels and the momentum factor $\beta$ used to preserve the past threshold information. For FADA~\cite{fada}, we consider the weight of the adversarial loss $\lambda$ and the temperature factor $T$ for feature scaling. The readers can refer to the original papers for more details.

Specifying a reasonable range for each hyper-parameter, we evaluate the methods' performance in a grid-search manner. The results confirm that these methods are more or less sensitive to hyper-parameter's selection. The maximum and minimum performance values in mean IOU are as follows; IntraDA: 44.0 vs. 32.4, FADA: 46.8 vs. 30.0, AdaptSegNet: 42.2 vs. 10.1, IAST: 50.7 vs. 43.6, AdvEnt: 43.7 vs. 6.9, and CBST: 46.9 vs. 45.3. More details are found in the supplementary material. 

\subsubsection{Selecting Hyper-parameters Using Few Validation Samples}

We next consider RQ2, i.e., how many labeled samples are necessary to find sufficiently good hyper-parameters. We conduct the  experiment as follows. For each method, we consider a set of $M$ different hyper-parameter settings on the grid in their ranges specified in the above experiment. We then perform UDA with each settings, resulting in $M$ models. We then evaluate these $M$ models
on the validation sets with different sizes, i.e., $\mathcal{S}_\mathrm{val}$'s with sizes $|\mathcal{S}_\mathrm{val}|=1,2,5,10,50$, and $100$, to pick the best performing one with each $\mathcal{S}_\mathrm{val}$, recording its performance on the test set $\mathcal{S}_\mathrm{test}$.

Table \ref{tbl:UDAvsVal} shows the results, in which we use the ratio of the average accuracy with the best hyper-parameters chosen using each of the $\mathcal{S}_\mathrm{val}$'s to that with the optimal hyper-parameters. In other words, these ratios show the deterioration of accuracy due to insufficient validation data. 

We can observe from Table \ref{tbl:UDAvsVal} 
that i) these UDA methods need more than a certain number of validation images to ensure that at least near optimal hyper-parameters are selected; using insufficient validation samples leads to large variance in performance; ii) different methods show different sensitivity to hyper-parameters; some methods are sensitive, whereas others are not. These can be visually confirmed in Fig.~\ref{fig:plotwshade}, which will be explained below. 


\begin{table*}
\centering
\begin{tabular}{l|cccccc} \hline 
$|\mathcal{S}_{\mathrm{val}}|$ & 1 & 2 & 5 & 10 & 50 & 100\\ \hline
IntraDA~\cite{intrada} & 98.9$\pm$3.5 & 99.3$\pm$2.4 & 99.6$\pm$1.7 & 99.6$\pm$1.5 & 99.8$\pm$1.4 & 100$\pm$1.0 \\ \hline
AdaptSegNet~\cite{adaptsegnet} & 95.8$\pm$1.9 & 96.8$\pm$1.8 & 98.8$\pm$1.0 & 99.3$\pm$0.6 & 99.8$\pm$0.5 & 100$\pm$0.3 \\ \hline
AdvEnt~\cite{advent} & 95.6$\pm$1.8 & 96.7$\pm$1.5 & 97.9$\pm$1.0 & 98.4$\pm$0.8 & 99.8$\pm$0.6 & 100$\pm$0.3 \\ \hline
CBST~\cite{cbst} & 98.5$\pm$0.6 & 98.1$\pm$0.6 & 98.5$\pm$0.6 & 98.9$\pm$0.8 & 99.8$\pm$0.6 & 100$\pm$0.5 \\ \hline
IAST~\cite{iast} & 98.0$\pm$1.6 & 99.0$\pm$1.3 & 99.2$\pm$0.5 & 99.4$\pm$0.4 & 99.8$\pm$0.4 & 100$\pm$0.4 \\ \hline
FADA~\cite{fada} & 98.5$\pm$0.8 & 98.9$\pm$0.9 & 98.9$\pm$1.0 & 99.8$\pm$0.8 & 100$\pm$0.6 & 100$\pm$0.5 \\ \hline
\end{tabular}
\caption{Ratio (\%) of performance (in mean IOU) of different UDA methods with the best hyper-parameters selected using $\mathcal{S}_\mathrm{val}$ to their highest accuracy.}
\label{tbl:UDAvsVal}
\end{table*}

\subsubsection{Comparison of UDA Methods and Finetuning}

We next consider RQ3, i.e., how effective simple finetuning is. Specifically, we finetune the model that is pre-trained on the source-domain data. Note that this is the same pre-trained model as that used in the UDA methods. We finetune it using the labeled target-domain samples in $\mathcal{S}_\mathrm{val}$ and repeat weight updates for a sufficiently large number, 2,000\footnote{We report the results of a more rigorous procedure in the supplementary material, i.e., splitting $\mathcal{S}_\mathrm{val}$ into two sets and using one for fine-tuning and the other for determining the iteration of weight updates.}. Using $\mathcal{S}_\mathrm{val}$ with different sizes $|\mathcal{S}_\mathrm{val}|\in\{1, 2, 5, 10, 50, 100\}$, we compare finetuning and UDA methods; finetuning uses $\mathcal{S}_\mathrm{val}$ for training and UDA methods use $\mathcal{S}_\mathrm{val}$ for obtaining the best hyper-parameters. We consider two adaptation scenarios, GTA5 $\rightarrow$ Cityscapes and SYNTHIA $\rightarrow$ Cityscapes.

\begin{figure*}[t]
\centering
\includegraphics[width=0.80\linewidth]{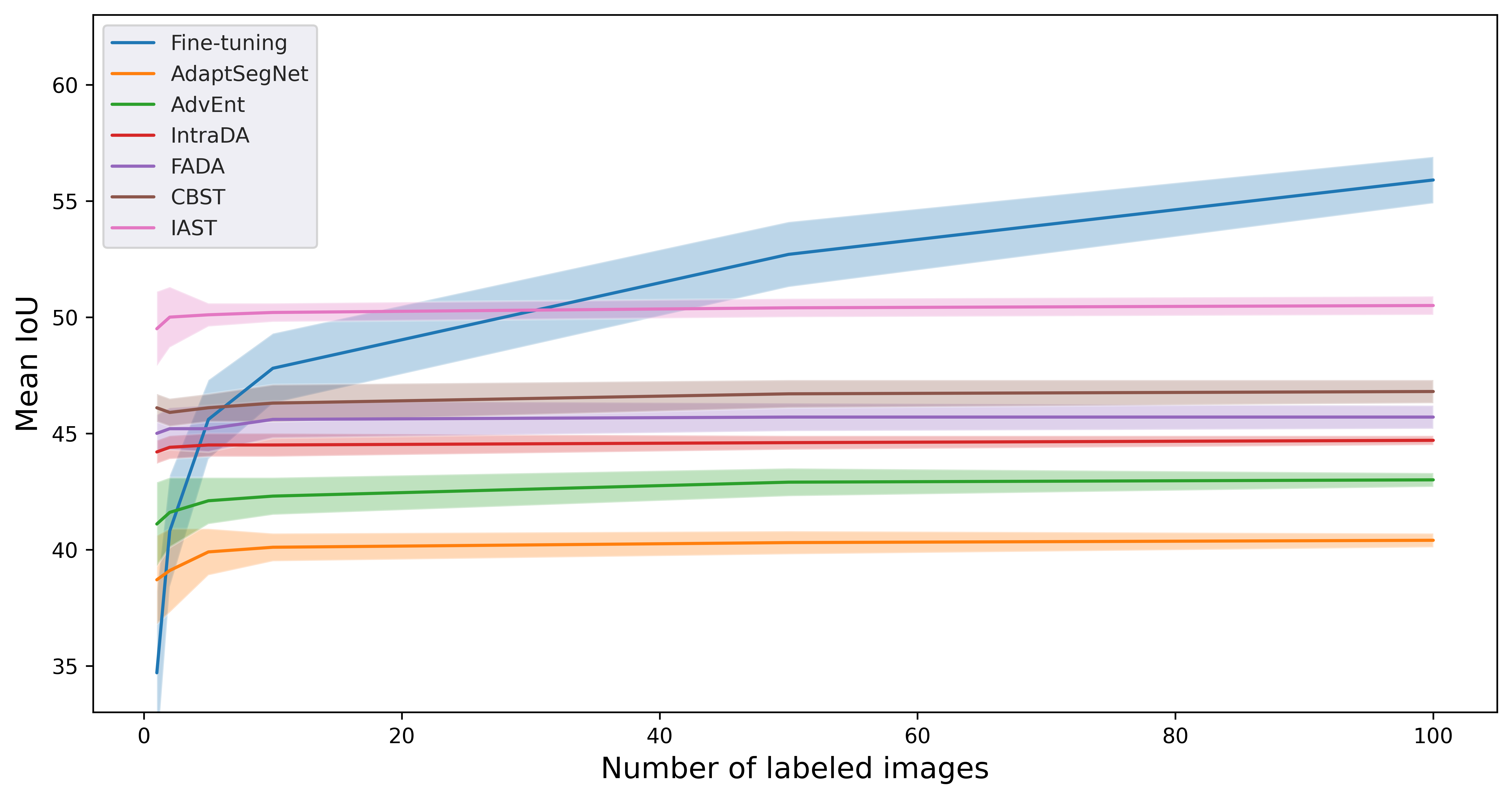}
\includegraphics[width=0.80\linewidth]{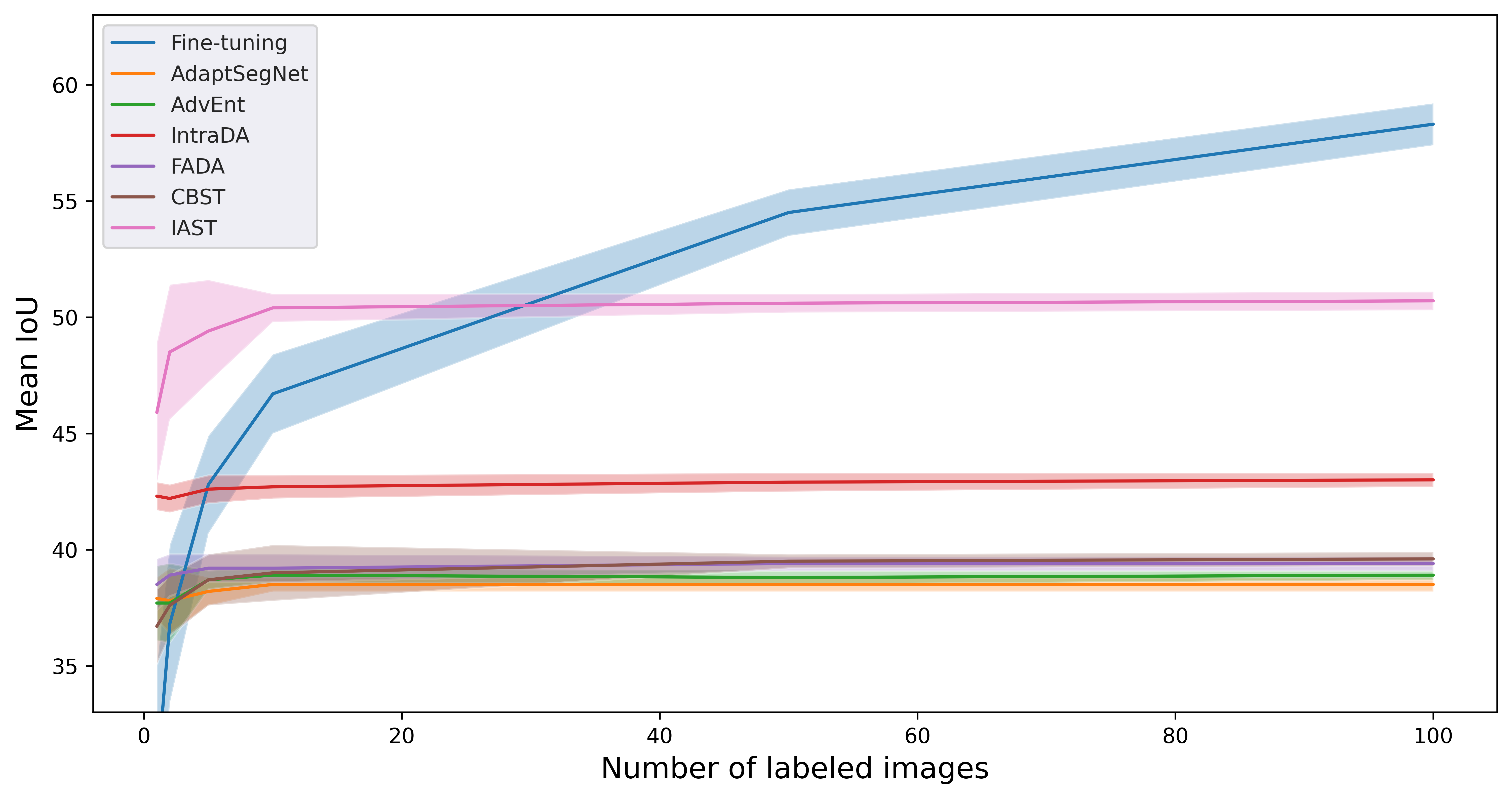}
\caption{Performance (mean IoU) of UDA methods and finetuning vs. the number of labeled images (i.e., $|\mathcal{S}_\mathrm{val}|$). Upper: GTA5 $\rightarrow$ Cityscapes. Lower: SYNTHIA $\rightarrow$ Cityscapes.}
\label{fig:plotwshade}
\end{figure*}

Figure \ref{fig:plotwshade} show the results; the upper plot shows GTA5 $\rightarrow$ Cityscapes and the lower one SYNTHIA $\rightarrow$ Cityscapes. We can observe the following. Finetuning using a few training images (i.e., $|\mathcal{S}_\mathrm{val}|=1$ or 2) performs worse than the UDA counterparts. It has a larger variance overall than UDA methods for a wide range of the number of images. However, as the number of images increases, finetuning starts to perform better than the UDA methods; it achieves on par with the second-best UDA method at $|\mathcal{S}_\mathrm{val}|=5$. IAST is much better than the other UDA methods, for which finetuning needs about 40 images to be better. We may conlclude that finetuning will be the first choice if we have more than 50 images for the considered scenarios. Overall, the results show the surprisingly strong performance of finetuning as a baseline method for adapting a model to a new domain.




%% file: sections/5_summary.tex
\section{Summary and Conclusion}

In this paper, we have reconsidered unsupervised domain adaptation (UDA) for semantic segmentation from a data-centric perspective. We first questioned the basic assumption of UDA that we have no access to labeled target domain samples, pointing out the contradiction with the standard procedure of ML that verifies the model's performance before its deployment. Next, assuming a small amount of labeled target-domain samples that are slightly more than necessary for the validation, we considered how we utilize them for better inference. Specifically, we first examined how many labeled samples are necessary and sufficient for selecting good hyper-parameters of each previous UDA method. We then examined finetuning using the same number of labeled samples. Finally, we compared the two, i.e., UDA methods and the simple finetuning in an equal condition in the number of labeled samples. 

The experimental results lead to the following findings. Firstly, different UDA methods show different sensitivity to the choice of hyper-parameters; while some UDA methods need only a small number of samples (i.e., images), as few as five, to choose the best hyper-parameters, this is not the case with other methods. Secondly, the simple finetuning works surprisingly well even with labeled samples of less than 100. 

These results imply that future studies on UDA must consider finetuning as a baseline to compare. In particular, they must first examine how many labeled samples are necessary to adjust the hyper-parameters of the proposed method and then show that it is superior to finetuning using the same (amount of) labeled samples. While this study considers the specific adaptation scenarios GTA5/SYNTHIA $\rightarrow$ Cityscapes, we think the conclusion will hold for other scenarios. Recall that our finetuning method uses only labeled target-domain samples; it leaves unlabeled target-domain samples untouched, which UDA methods utilize. Using them additionally will undoubtedly lead to better performance. However, it is unclear how to do it, where we need to consider a smaller number of labeled samples than in previous SSDA studies, as mentioned earlier. We will study this in the future.

%% file: sections/X_supplementary.tex
\appendix
{\Large
\textbf{Appendix}
}

\section{Detailed Results of UDA Methods with Different Hyper-parameters}

\begin{figure}[h]
\centering
\includegraphics[width=60mm]{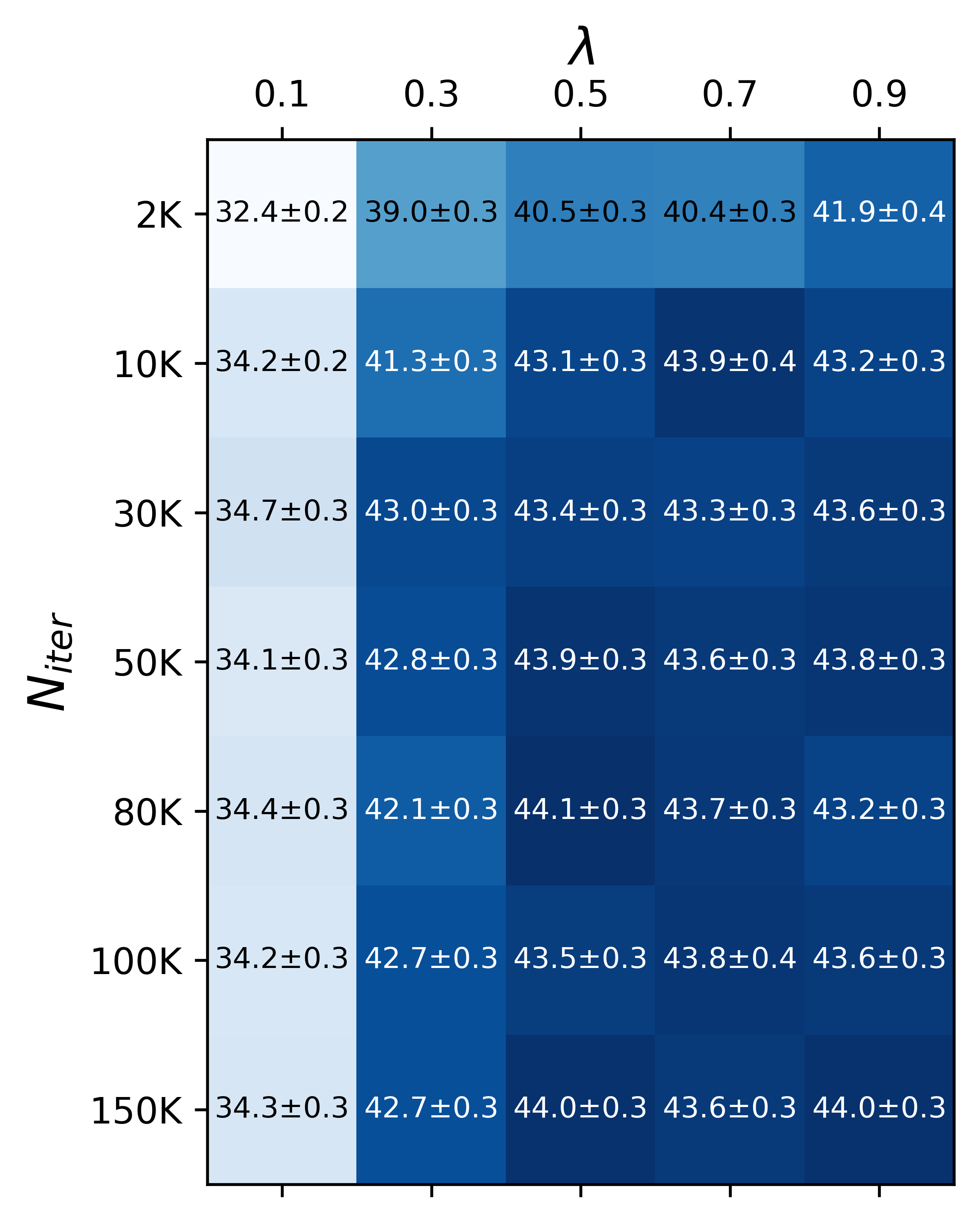}
\includegraphics[width=60mm]{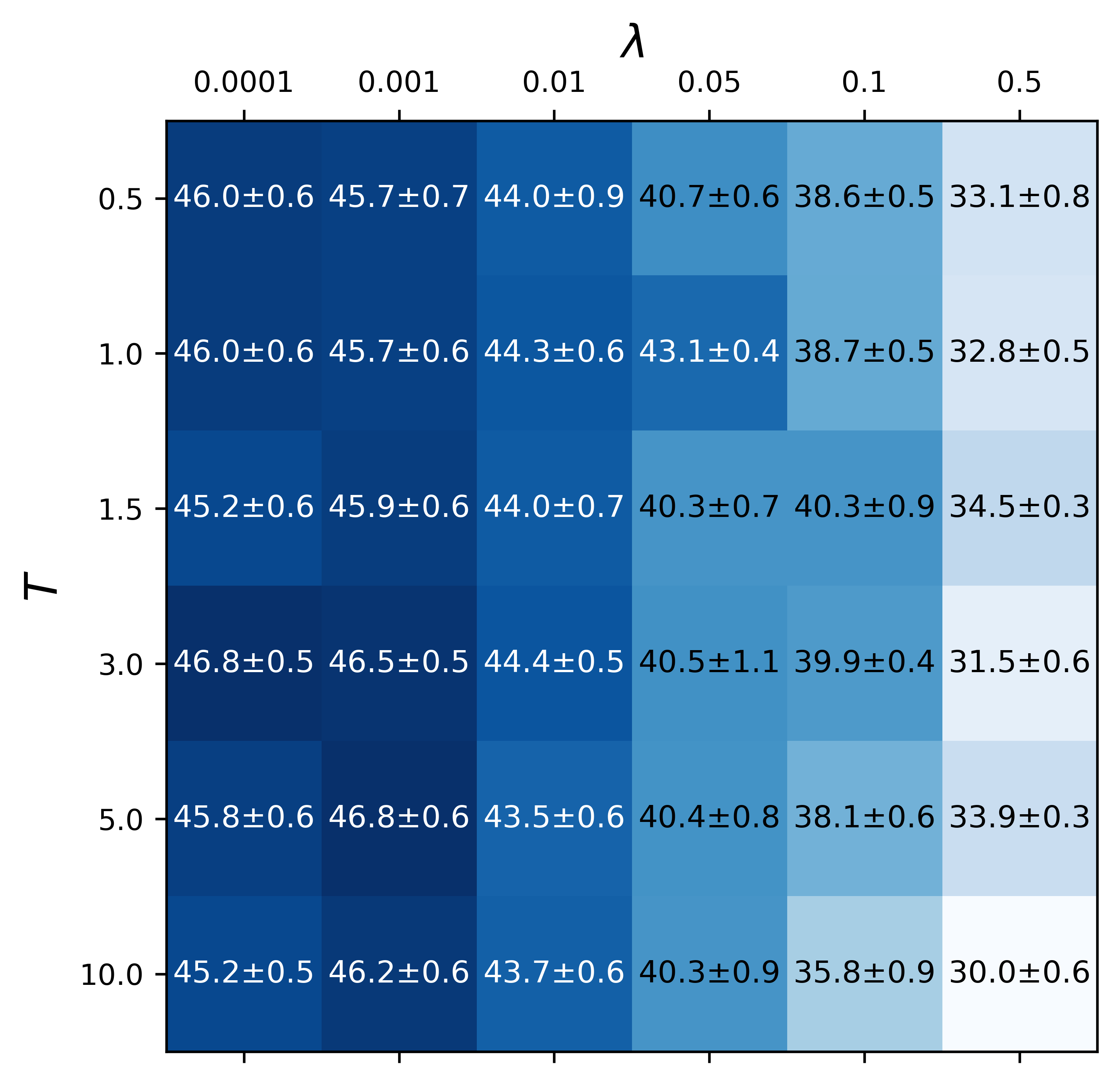}
\caption{Hyper-parameter sensitivity of IntraDA (upper) and FADA (lower).}
\label{fig:sen_fig1}
\end{figure}

\begin{figure}[h]
\centering
\includegraphics[width=60mm]{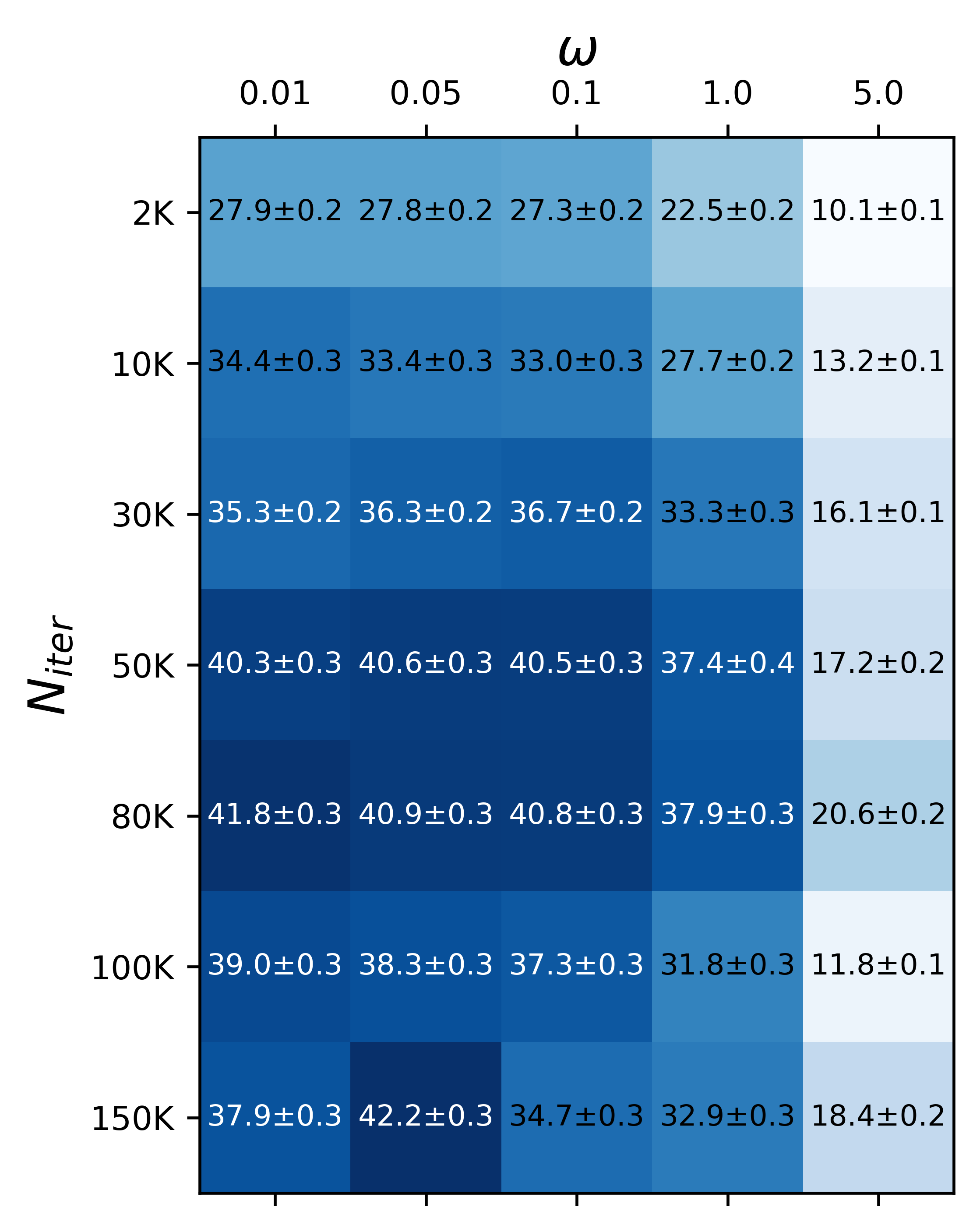}
\includegraphics[width=60mm]{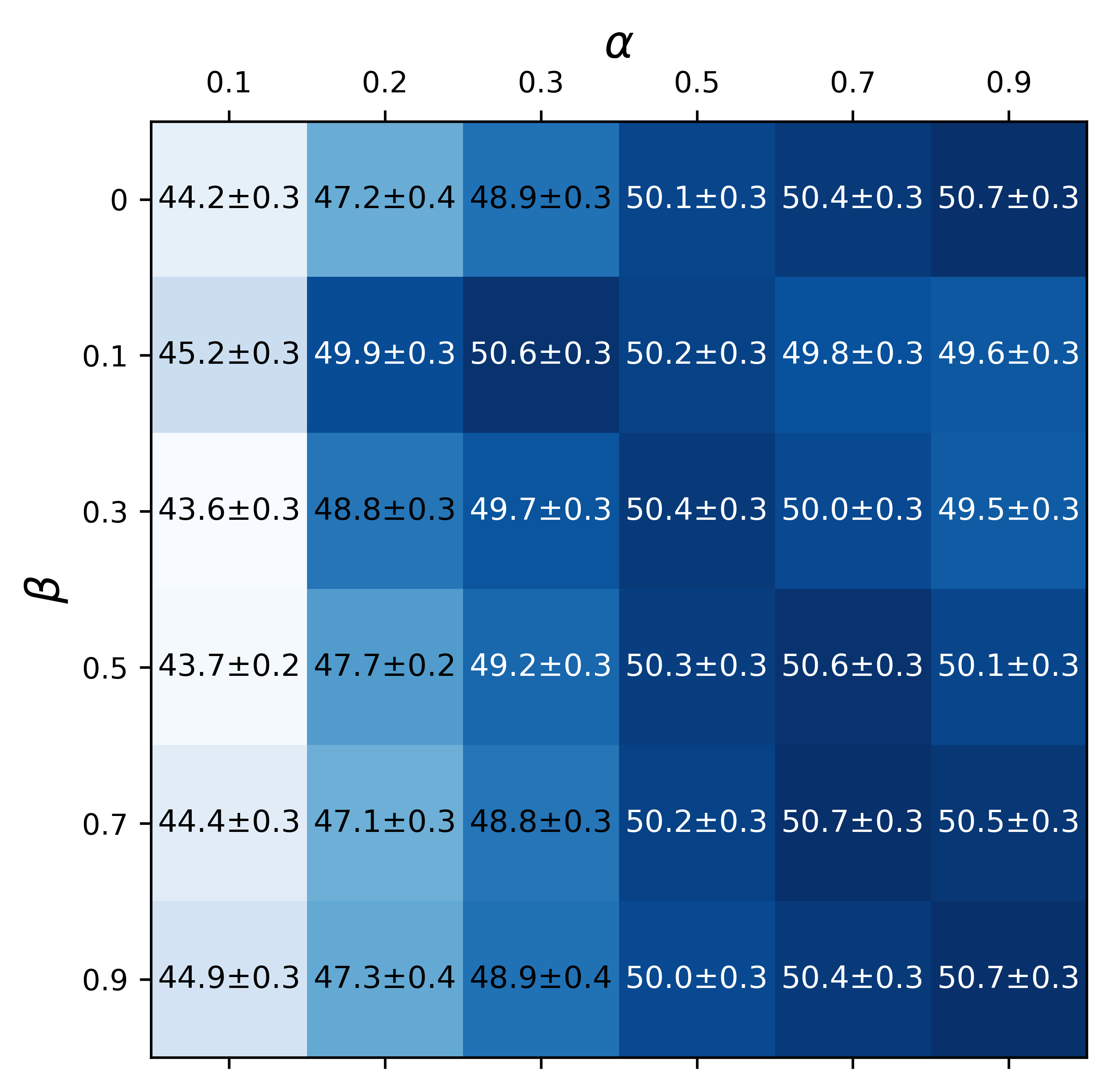}
\caption{Hyper-parameter sensitivity of AdaptSegNet (upper) and IAST (lower).}
\label{fig:sen_fig2}
\end{figure}

\begin{figure}[h]
\centering
\includegraphics[width=60mm]{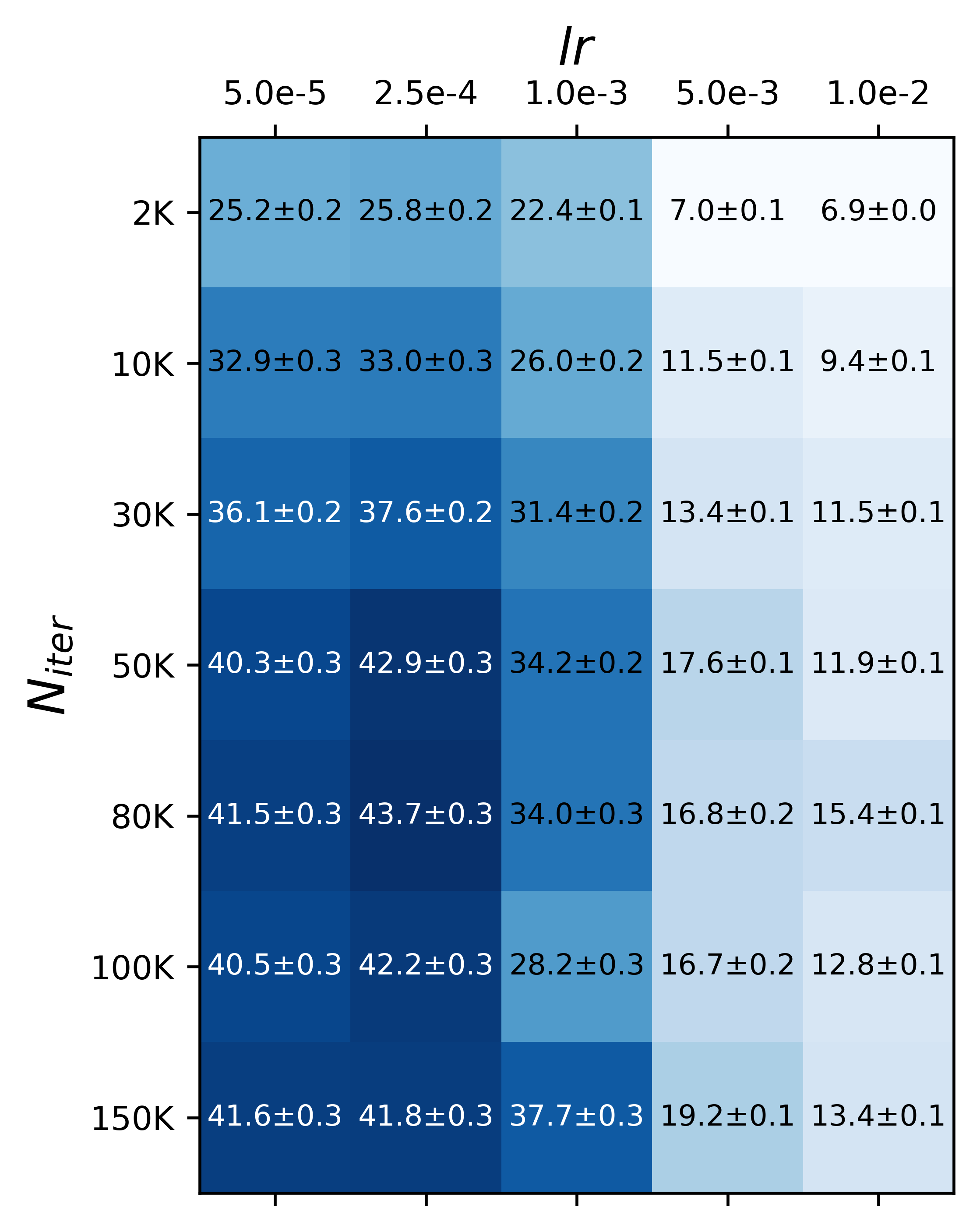}
\includegraphics[width=60mm]{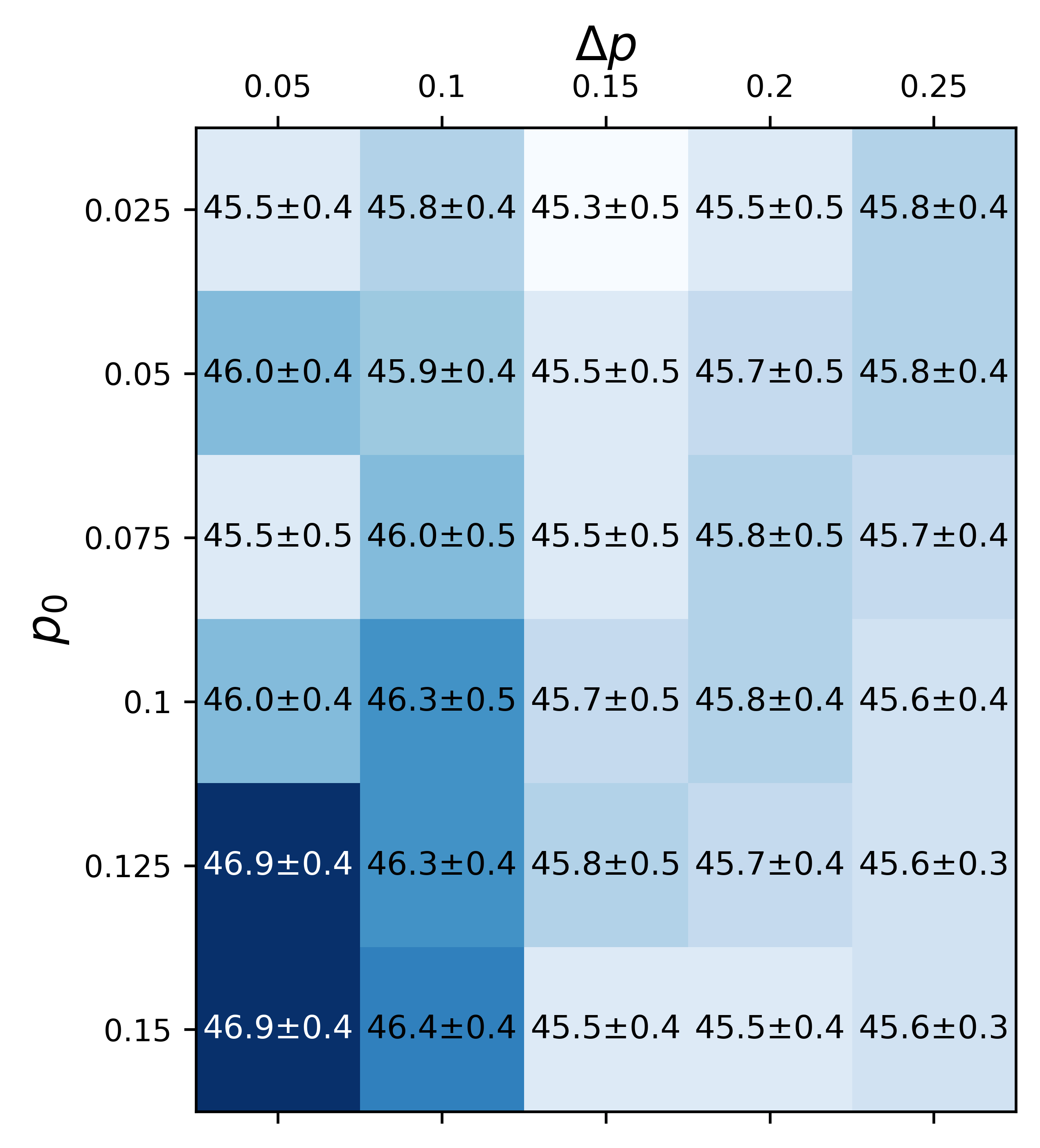}
\caption{Hyper-parameter sensitivity of AdvEnt (upper) and CBST (lower).}
\label{fig:sen_fig3}
\end{figure}

Figure~\ref{fig:sen_fig1}, \ref{fig:sen_fig2}, and \ref{fig:sen_fig3} show the full results of the six UDA methods with different hyper-parameter settings for the experiment described in Sec. 4.2.1 in the paper.
The numbers in each grid represent the mean IoU value and standard deviation by the corresponding hyper-parameter combinations obtained from 30 trials; darker colors show better performance.

\section{More Rigorous Evaluation of Finetuning}

To avoid further complication, in the experiments reported in the main paper, we fix the number of iterations for finetuning to a sufficiently large number, 2K. This configuration simulates the case where finetuning stably converges to a solution as the number of epochs grows (to infinity). However, this is arguably unfair in a strict sense; we may need to conduct a validation step, in which we use a held-out set of labeled target domain data to select the best epochs, as with UDA. Thus, we tested this in a supplementary experiment; specifically, instead of using the entire $\mathcal{S}_\mathrm{val}$ for training (i.e., finetuning), we split it into two without overlap, as $\mathcal{S}_\mathrm{val}=\mathcal{S}^\mathrm{FT}_\mathrm{train}\cup\mathcal{S}^\mathrm{FT}_\mathrm{val}$; we then finetune the model on $\mathcal{S}^\mathrm{FT}_\mathrm{train}$ and choose the best epoch using $ \mathcal{S}^\mathrm{FT}_\mathrm{val}$. The data splitting scheme is summarized as follows:

\medskip
\noindent
{\bf S2}: 
We initially follow the scheme in our main paper (Eq. 1). Namely, a set $\mathcal{S}_\mathrm{test}$ consisting of 400 samples is first chosen from the original validation set of Cityscapes. Then $\mathcal{S}_\mathrm{val}$ is chosen from the rest, which has a smaller size than 100. Next, we split it into $\mathcal{S}^\mathrm{FT}_\mathrm{train}$ and $\mathcal{S}^\mathrm{FT}_\mathrm{val}$ with different sizes such that $\mathcal{S}_\mathrm{val}=\mathcal{S}^\mathrm{FT}_\mathrm{train}\cup\mathcal{S}^\mathrm{FT}_\mathrm{val}$ and $\mathcal{S}^\mathrm{FT}_\mathrm{train}\cap\mathcal{S}^\mathrm{FT}_\mathrm{val}=\phi$.

\medskip
\noindent
A question is how to split $\mathcal{S}_\mathrm{val}$. We test several splits of $\mathcal{S}_\mathrm{val}$ in its critical range recognized in the main experiments, i.e., $|\mathcal{S}_\mathrm{val}|=\{10, 50, 100\}$.


The detailed setting of the experiments is as follows. For finetuning, we train the model with $\mathcal{S}^\mathrm{FT}_\mathrm{train}$ for 5K iterations, saving a checkpoint for every 500 iterations and select one with the best performance on $\mathcal{S}^\mathrm{FT}_\mathrm{val}$ for testing. For UDA methods, we employ the same procedure as before; their hyper-parameters are selected using  $\mathcal{S}_\mathrm{val}=\mathcal{S}^\mathrm{FT}_\mathrm{train} \cup \mathcal{S}^\mathrm{FT}_\mathrm{val}$. As in the previous experiments, we report the average and variance over trials  $i=1,\ldots,N_\mathrm{trial}$ with $N_\mathrm{trial}=30$. 

\begin{figure*}
\centering
\includegraphics[width=0.8\linewidth]{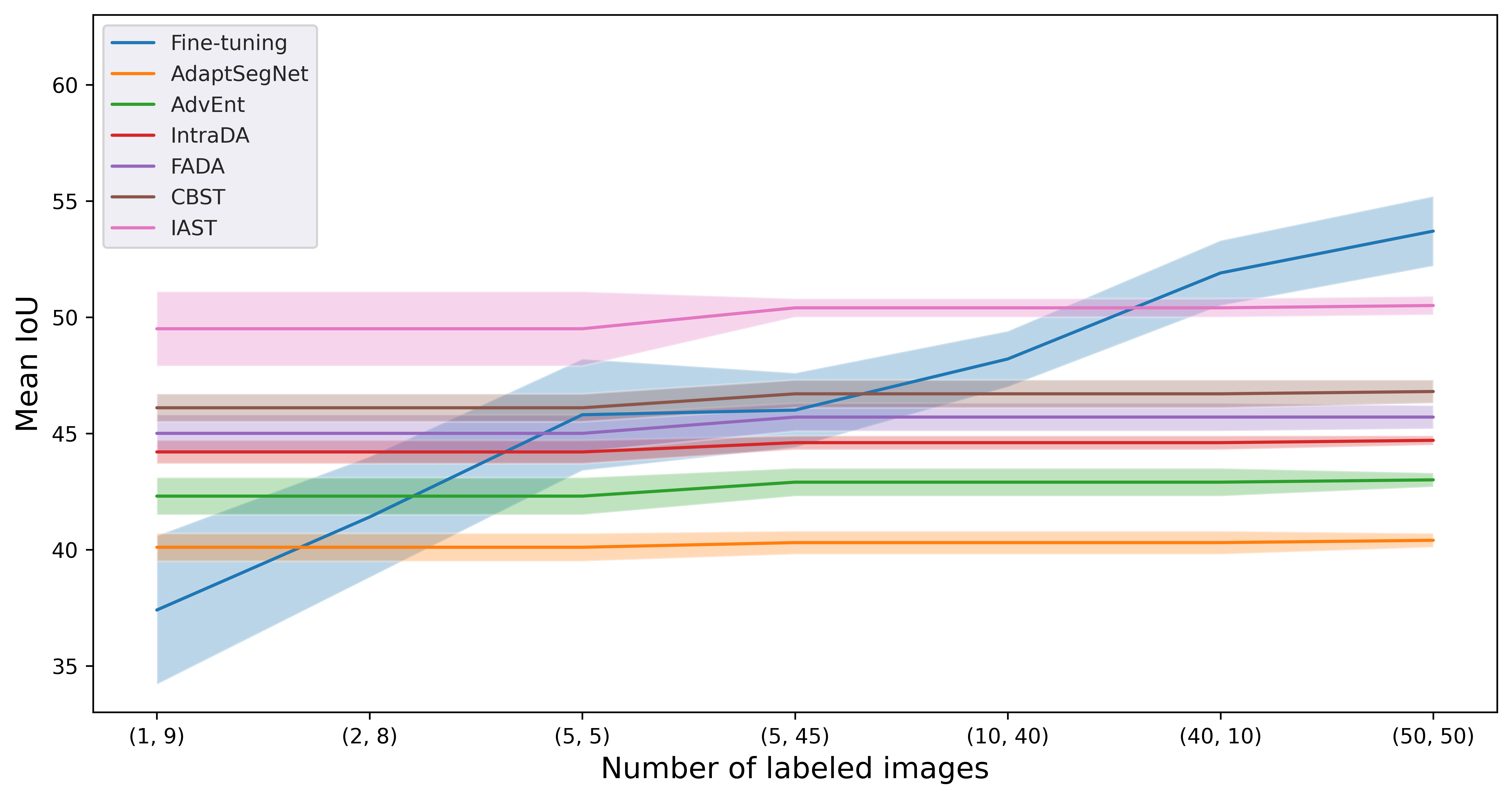}
\caption{Performance (mean IoU) of UDA methods and finetuning vs. the number of labeled images (i.e., ($|\mathcal{S}^\mathrm{FT}_\mathrm{train}|$, $|\mathcal{S}^\mathrm{FT}_\mathrm{val}|$)) on GTA5 $\rightarrow$ Cityscapes.}
\label{fig:plotwshade_supp}
\end{figure*}

Table~\ref{tbl:supp} and Figure~\ref{fig:plotwshade_supp} show the results. We can see that a similar observation to Sec. 4.2.3 holds. The finetuning method works better than the UDA methods when $\lvert\mathcal{S}_\mathrm{val}\lvert\geq 50$ (and $\lvert\mathcal{S}^\mathrm{FT}_\mathrm{train}\rvert\geq 40)$. In more detail, finetuning performs on par with UDA methods but IAST at $\lvert\mathcal{S}^\mathrm{FT}_\mathrm{train}\rvert = 5$ and better with $\geq 10$. It outperforms IAST at $\lvert\mathcal{S}^\mathrm{FT}_\mathrm{train}\rvert = 40$. While the overall performance variance of finetuning is larger when $|\mathcal{S}_\mathrm{val}^\mathrm{FT}|$ is very small (e.g., $=5$), the variance may be acceptable when $|\mathcal{S}_\mathrm{val}^\mathrm{FT}|\geq 10$.

\begin{table*}
\caption{Performance (mean IoU) of UDA methods and finetuning with scheme 2. GTA5 $\rightarrow$ Cityscapes}
\centering
\begin{tabular}{l|ccccccccccccc}
\hline
\rule{0pt}{2.6ex}
$(\lvert\mathcal{S}^\mathrm{FT}_\mathrm{train}\rvert, \lvert\mathcal{S}^\mathrm{FT}_\mathrm{val}\rvert)$ & (1, 9) & (2, 8) & (5, 5) & (5, 45) & (10, 40) & (40, 10) & (50, 50) \\ [0.6ex]
\hline
Finetuning & 37.4$\pm$3.2 & 41.4$\pm$2.6 & 45.8$\pm$2.4 & 46.0$\pm$1.6 & 48.2$\pm$1.2 & 51.9$\pm$1.4 & 53.7$\pm$1.5 \\
AdaptSegNet~\cite{adaptsegnet} & 40.1$\pm$0.6 & 40.1$\pm$0.6 & 40.1$\pm$0.6 & 40.3$\pm$0.5 & 40.3$\pm$0.5 & 40.3$\pm$0.5 & 40.4$\pm$0.3 \\
AdvEnt~\cite{advent} & 42.3$\pm$0.8 & 42.3$\pm$0.8 & 42.3$\pm$0.8 & 42.9$\pm$0.6 & 42.9$\pm$0.6 & 42.9$\pm$0.6 & 43.0$\pm$0.3 \\
IntraDA~\cite{intrada} & 44.2$\pm$0.5 & 44.2$\pm$0.5 & 44.2$\pm$0.5 & 44.6$\pm$0.3 & 44.6$\pm$0.3 & 44.6$\pm$0.3 & 44.7$\pm$0.2 \\
FADA~\cite{fada} & 45.0$\pm$0.8 & 45.0$\pm$0.8 & 45.0$\pm$0.8 & 45.7$\pm$0.6 & 45.7$\pm$0.6 & 45.7$\pm$0.6 & 45.7$\pm$0.5 \\
CBST~\cite{cbst} & 46.1$\pm$0.6 & 46.1$\pm$0.6 & 46.1$\pm$0.6 & 46.7$\pm$0.6 & 46.7$\pm$0.6 & 46.7$\pm$0.6 & 46.8$\pm$0.5 \\
IAST~\cite{iast} & 49.5$\pm$1.6 & 49.5$\pm$1.6 & 49.5$\pm$1.6 & 50.4$\pm$0.4 & 50.4$\pm$0.4 & 50.4$\pm$0.4 & 50.5$\pm$0.4 \\
\hline 
\end{tabular}
\label{tbl:supp}
\end{table*}

\section{Detailed Comparison of UDA Methods and Finetuning}

Although we provide the plots for the comparison of the UDA methods and finetuning in the main paper (i.e., Fig.~1), Table~\ref{tbl:finetune_g2c_s2} and \ref{tbl:finetune_s2c_s2} show their exact mean IoU and standard deviation values on GTA5 $\rightarrow$ Cityscapes and SYNTHIA $\rightarrow$ Cityscapes scenarios, respectively. 

\begin{table*}
\centering
\begin{tabular}{l|ccccccccc}
\hline
$|\mathcal{S}_{\mathrm{val}}|$ & 1 & 2 & 5 & 10 & 50 & 100 \\
\hline
Finetuning & 34.7$\pm$3.5 & 40.8$\pm$2.4 & 45.6$\pm$1.7 & 47.8$\pm$1.5 & 52.7$\pm$1.4 & 55.9$\pm$1.0 \\
AdaptSegNet~\cite{adaptsegnet} & 38.7$\pm$1.9 & 39.1$\pm$1.8 & 39.9$\pm$1.0 & 40.1$\pm$0.6 & 40.3$\pm$0.5 &40.4$\pm$0.3\\
AdvEnt~\cite{advent} & 41.1$\pm$1.8 & 41.6$\pm$1.5 & 42.1$\pm$1.0 & 42.3$\pm$0.8 & 42.9$\pm$0.6 & 43.0$\pm$0.3 \\
IntraDA~\cite{intrada} & 44.2$\pm$0.5 & 44.4$\pm$0.5 & 44.5$\pm$0.5 & 44.5$\pm$0.5 & 44.6$\pm$0.3 & 44.7$\pm$0.2 \\
FADA~\cite{fada} & 45.0$\pm$0.8 & 45.2$\pm$0.9 & 45.2$\pm$1.0 & 45.6$\pm$0.8 & 45.7$\pm$0.6 & 45.7$\pm$0.5 \\
CBST~\cite{cbst} & 46.1$\pm$0.6 & 45.9$\pm$0.6 & 46.1$\pm$0.6 & 46.3$\pm$0.8 & 46.7$\pm$0.6 & 46.8$\pm$0.5 \\
IAST~\cite{iast} & 49.5$\pm$1.6 & 50.0$\pm$1.3 & 50.1$\pm$0.5 & 50.2$\pm$0.4 & 50.4$\pm$0.4 & 50.5$\pm$0.4 \\
\hline
\end{tabular}
\caption{Performance (mean IoU) of UDA methods and finetuning on GTA5 $\rightarrow$ Cityscapes}
\label{tbl:finetune_g2c_s2}
\end{table*}

\begin{table*}
\centering
\begin{tabular}{l|ccccccccc}
\hline
$|\mathcal{S}_{\mathrm{val}}|$ & 1 & 2 & 5 & 10 & 50 & 100 \\
\hline
Finetuning & 30.1$\pm$4.8 & 36.8$\pm$3.4 & 42.8$\pm$2.1 & 46.7$\pm$1.7 & 54.5$\pm$1.0 & 58.3$\pm$0.9 \\
AdaptSegNet~\cite{adaptsegnet} & 37.9$\pm$0.9 & 37.8$\pm$1.4 & 38.2$\pm$0.6 & 38.5$\pm$0.3 & 38.5$\pm$0.3 & 38.5$\pm$0.3 \\
AdvEnt~\cite{advent} & 37.7$\pm$1.6 & 37.7$\pm$1.7 & 38.7$\pm$0.4 & 38.9$\pm$0.3 & 38.8$\pm$0.3 & 38.9$\pm$0.2 \\
IntraDA~\cite{intrada} & 42.3$\pm$0.6 & 42.2$\pm$0.6 & 42.6$\pm$0.6 & 42.7$\pm$0.5 & 42.9$\pm$0.4 & 43.0$\pm$0.3 \\
FADA~\cite{fada} & 38.5$\pm$1.1 & 38.9$\pm$0.9 & 39.2$\pm$0.6 & 39.2$\pm$0.6 & 39.4$\pm$0.3 & 39.4$\pm$0.3 \\
CBST~\cite{cbst} & 36.7$\pm$1.6 & 37.6$\pm$1.3 & 38.7$\pm$1.1 & 39.0$\pm$1.2 & 39.5$\pm$0.3 & 39.6$\pm$0.3 \\
IAST~\cite{iast} & 45.9$\pm$3.0 & 48.5$\pm$2.9 & 49.4$\pm$2.2 & 50.4$\pm$0.6 & 50.6$\pm$0.4 & 50.7$\pm$0.4 \\
\hline
\end{tabular}
\caption{Performance (mean IoU) of UDA methods and finetuning on SYNTHIA $\rightarrow$ Cityscapes}
\label{tbl:finetune_s2c_s2}
\end{table*}